\documentclass[10pt,twocolumn,letterpaper]{article}

\usepackage[pagenumbers]{cvpr} 

\definecolor{cvprblue}{rgb}{0.21,0.49,0.74}
\usepackage[pagebackref,breaklinks,colorlinks,allcolors=cvprblue]{hyperref}

\def\paperID{13762} 
\def\confName{CVPR}
\def\confYear{2025}

\title{Boltzmann Attention Sampling for Image Analysis with Small Objects}

\author{Theodore Zhao\thanks{Equal contribution.} \hspace{1cm}
Sid Kiblawi\footnotemark[1] \hspace{1cm}
Naoto Usuyama \hspace{1cm}
Ho Hin Lee \hspace{1cm} \\
Sam Preston \hspace{1cm} 
Hoifung Poon \hspace{1cm}
Mu Wei\thanks{Corresponding authors.} \hspace{1cm} 
\\
\begin{tabular}{c}
        Microsoft, Redmond, WA, USA \\
\url{https://aka.ms/boltzformer}
    \end{tabular}
}

\def\ourmethod{\textit{BoltzFormer}\xspace}

\begin{document}
\maketitle
\begin{abstract}

Detecting and segmenting small objects, such as lung nodules and tumor lesions, remains a critical challenge in image analysis. These objects often occupy less than 0.1\% of an image, making traditional transformer architectures inefficient and prone to performance degradation due to redundant attention computations on irrelevant regions. Existing sparse attention mechanisms rely on rigid hierarchical structures, which are poorly suited for detecting small, variable, and uncertain object locations.
In this paper, we propose {\em \ourmethod}, a novel transformer-based architecture designed to address these challenges through dynamic sparse attention. \ourmethod identifies and focuses attention on relevant areas by modeling uncertainty using a Boltzmann distribution with an annealing schedule. Initially, a higher temperature allows broader area sampling in early layers, when object location uncertainty is greatest. As the temperature decreases in later layers, attention becomes more focused, enhancing efficiency and accuracy.
\ourmethod seamlessly integrates into existing transformer architectures via a modular Boltzmann attention sampling mechanism. Comprehensive evaluations on benchmark datasets demonstrate that \ourmethod significantly improves segmentation performance for small objects while reducing attention computation by an order of magnitude compared to previous state-of-the-art methods.

\end{abstract}

\section{Introduction}
\label{sec:intro}

Transformer~\cite{vaswani2017attention} has significantly changed the field of image analysis. Its attention-based architecture introduces flexibility in handling inputs from various modalities, including images, text, and other forms of prompts. Seminal works such as SAM \cite{kirillov2023segment}, SAM 2 \cite{ravi2024sam}, and SEEM \cite{zou2024segment} unify segmentation tasks for various object types, making the models flexible and generalizable. There is also rising interest in adapting such promptable segmentation models to key application domains, such as biomedicine (e.g., MedSAM \cite{ma2024segment}, BiomedParse \cite{zhao2024biomedparse}, and MedSAM-2 \cite{zhu2024medical}). \\


While the progress is exciting, challenges abound. A salient growth area is small objects, which are particularly difficult for conventional segmentation models, especially when there is no user-provided bounding box and only the object description is available as input~\cite{cheng2022masked, zhao2024biomedparse}. This challenge is especially pronounced in critical application domains such as biomedicine, when important objects like lung nodules, tumor lesions, vessels, and many other anatomical structures are tiny, sometimes occupying less than 0.1\% of the image. 
By default, many standard segmentation models require the user to conduct the object detection step and provide object-specific spatial prompts as input, such as bounding boxes, points, or scribbles~\cite{ma2024segment,wong2024scribbleprompt,zhu2024medical,lee2024foundation}. In many applications, there are a large number of objects and this approach is clearly not scalable. 
Ideally, a user can simply specify the object description in a text prompt, and the model can automatically detect and segment the given objects, as in BiomedParse~\cite{zhao2024biomedparse}. However, in such an end-to-end setting, small objects pose significant challenges because they are particularly hard to locate and identify.



The difficulty in detecting and segmenting small objects originates from the nature of the attention mechanism in transformer. For a small object, the relevant information is confined within a tiny portion of the image (the object and its neighborhood). This means that the vast majority of attention computation would be spent on irrelevant pixels far away from the object. In addition to being wasteful, this also introduces significant distractions and noises, making the learning less efficient. Prior work has attempted to address this issue by focusing decoding on the foreground regions \cite{zhang2023mp, cheng2022masked}. However, they used relatively rigid rules, which are ill-suited for small objects due to their variability and the uncertainty about their locations. \\



\begin{figure*}[ht]
    \centering
    \includegraphics[trim={0 7.5cm 0 0},clip, width=0.95\linewidth]{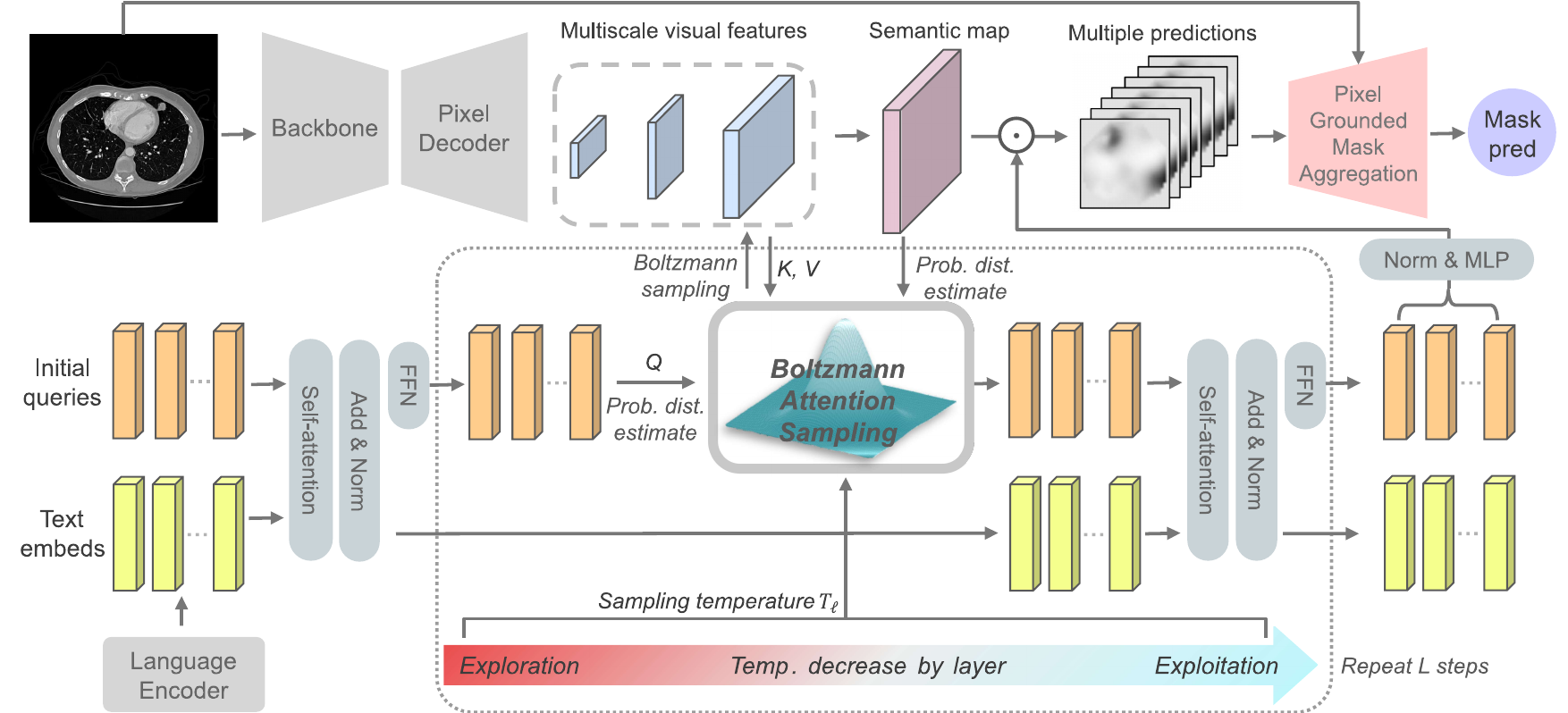}
    \caption{\small 
    The overall architecture of \ourmethod for end-to-end object detection and segmentation via a unified text prompt. \ourmethod is a novel transform-based architecture that introduces a Boltzmann attention sampling module to dynamically propose sparse areas to focus cross-attention in each layer using a Boltzmann distribution. To account for uncertainty, which is especially high in earlier stage of computation, \ourmethod starts with a high temperature in the first layer, which gradually cools down in subsequent layers. This is reminiscent of a reinforcement learning process, where exploration is favored in the initial layers (more sparse areas being sampled), and exploitation in later layers (focusing on a handful of most promising areas).    
    The model takes image (upper left) and text prompt (lower left) as input, and outputs segmentation mask (upper right) for the object specified in the text prompt.     
    Specifically, we use a standard image encoder to obtain multiscale visual features, including a high-resolution semantic map (upper middle). \ourmethod starts with a set of latent queries that try to model the correct semantic for the prompted object in the image (middle left). 
    In each layer, the query vectors are combined with the semantic map to produce a Boltzmann distribution over the image, which is then used to sample the sparse areas. They each attend exclusively to the visual features in the sampled area and update themselves (see \autoref{fig:attention} for more details). The queries communicate with the text embeddings through self-attention after each Boltzmann attention sampling layer (center block).  
    After the transformer layers, each query is combined with the image semantic map to generate a candidate predicted mask. The predictions are aggregated by a pixel grounded mask aggregation (PiGMA) module into the final mask prediction (upper right, see \autoref{fig:pixel} for details). 
    }
    \label{fig:model}
\end{figure*}

In this paper, we propose a novel transformer architecture, \ourmethod, which learns to use a Boltzmann distribution to dynamically propose sparse candidate areas to focus attention computation on. \ourmethod takes an image and a natural-language description as input and conducts end-to-end detection and segmentation of the object as specified by the text prompt, as illustrated in Fig. \ref{fig:model}.
To accommodate uncertainty about the object location, \ourmethod employs an annealing schedule, which starts with a high temperature for the Boltzmann distribution so that it would propose more areas to sample when the uncertainty is the highest, and cools down in subsequent layers as the object location becomes clearer so that the model can focus on identifying the details in the object segment.
The Boltzmann sampling step is modular and can be easily incorporated into any modern transformer-based image analysis process. 

We develop a reference system for \ourmethod by combining the Boltzmann sampling module with state-of-the-art design for end-to-end image analysis. Specifically, we use a state-of-the-art image encoder to extract multi-scale visual features and a semantic map, and learn a set of latent query vectors in tandem with the image representation and text embedding. 
The learning is unrolled in a deep architecture with transformer-based layers augmented by the Boltzmann attention sampling module.
In the final layer, the query vectors are combined with the image representation to generate candidate segmentation masks, which are aggregated by a Pixel Grounded Mask Aggregation (PiGMA) module to produce the final segmentation mask output. 

We highlight our contributions as follows:
\begin{itemize}
    \item To the best of our knowledge, we are the first to incorporate Boltzmann sampling into transformer to focus attention computation on sparse areas of the image that most likely contain the queried object. We conducted thorough ablation studies to establish the best practice for this novel transformer architecture.
    \item We propose a novel PiGMA module to efficiently aggregate an ensemble of segmentation mask predictions based on the query vectors. Through ablation studies, we demonstrate the effectiveness of this ensemble approach.
    \item We conducted extensive evaluation on end-to-end object segmentation via a text prompt, using challenging segmentation datasets with objects of size ranging from 0.002\%-20\% of the image. 
    Remarkably, \ourmethod substantially outperforms prior best methods, attaining a gain of 3-12 absolute points in mean Dice score, while reducing attention computation by an order of magnitude.
\end{itemize}

\section{Related work}

Segmentation and detection have been the core parts of image analysis, especially in medical images. Architecture wise, the field witnesses the progress from convolutional neural networks (CNNs), particularly architectures like Fast R-CNN~\cite{girshick2015fast}, {U-Net}~\cite{ronneberger2015u}, {Mask R-CNN}~\cite{he2017mask} and Sparse R-CNN \cite{sun2021sparse} to recent development of transformers. This transformer revolution led to seminal works including DETR~\cite{carion2020end}, MaskFormer~\cite{cheng2021per}, and Mask2Former~\cite{cheng2022masked}. All these works used transformer decoder as the unifying architecture.

These transformer decoder for segmentation and detection nourished the recent developments toward generalized promptable image segmentation models, including Segment Anything Model (SAM)~\cite{kirillov2023segment}, SAM 2~\cite{ravi2024sam}, and Segment Everything Everywhere Model (SEEM)~\cite{zou2024segment}. All these models feature a transformer mask decoder with flexibility in taking prompts such as points, bounding boxes and text. While the image encoder and prompt encoder are flexibly chosen, the transformer mask decoder plays the critical role in performing quality segmentation. The decoders of SAM and SAM 2 were derived from MaskFormer~\cite{cheng2021per}, and SEEM was built upon the {Mask2Former}~\cite{cheng2022masked} architecture. Our work aims to improve upon the above mentioned transformer mask decoder architectures in the text prompting scenario.




\begin{figure}[t]
    \centering
    \includegraphics[trim={0 0 0 0},clip, width=\linewidth]{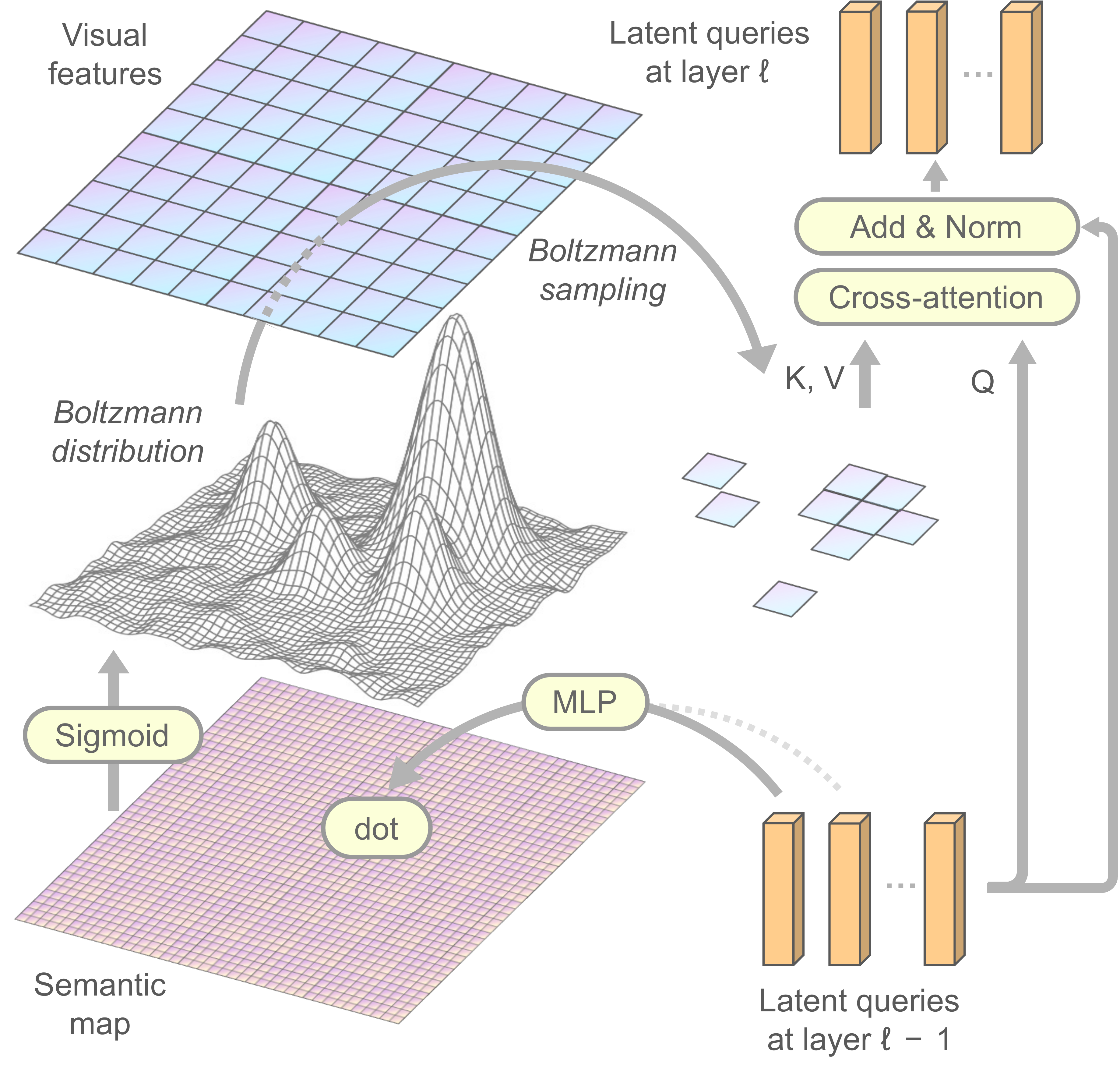}
    \caption{\small Illustration of the Boltzmann attention sampling block (center block in Fig. \ref{fig:model}). The latent queries from the previous layer each goes through the MLP transformation (Eq. \eqref{eq-mlp}) with dimension kept constant. Each transformed query vector takes dot product with all feature vectors on the semantic map, yielding scalars on the map. We use sigmoid to transform the scores into (0,1), and compute the Boltzmann distribution of temperature $\tau_\ell$ using Eq. \eqref{eq-boltz}. We then draw from the distribution to sample the corresponding patches in the visual feature for $N$ trials with replacement. The query attends exclusively to the samples features and add to itself. We perform the same for all query vectors and apply layer normalization on them at the end.}
    \label{fig:attention}
\end{figure}

There have been several advances in biomedical segmentation by utilizing the architectures presented above and training them with more biological focused data. MedSAM~\cite{ma2024segment} was finetuned on SAM using a large-scale dataset of over 1.5 million image-mask pairs across 10 imaging modalities and 30 cancer types, demonstrating superior segmentation performance and adaptability compared to modality-specific models. {MedSAM 2}~\cite{ma2024segmentmedicalimagesvideos} extends MedSAM to 3D medical imaging, supporting segmentation in volumetric and time-sequential data. BiomedParse~\cite{zhao2024biomedparse}, built upon SEEM, is a biomedical foundational model in that unifies segmentation, detection, and recognition tasks. By using text prompts, BiomedParse removes manual bounding-box interactions and enables scalable and accurate segmentation. In evaluations, BiomedParse outperformed state-of-the-art models, particularly for objects with complex and irregular shapes. However, finding small objects with text prompt is still a challenge.

In order to improve segmentation performance, researchers have developed attention masking techniques. One closely approach is in Mask2Former~\cite{cheng2022masked}, where the masked attention operation is hard thresholded by the previous layer's prediction. However, the prediction is not consistent across the layers, and the model struggles with small objects. MP-Former~\cite{zhang2023mp} proposed to introduce ground truth mask with point noise during training, however, the distribution difference between training and inference amplifies the inference time error, which is severe for small objects. 
In contrast, the proposed Boltzmann attention sampling uses probabilistic sampling to promote identifying the right regions throughout all layers. We compared our Boltzmann sampling strategy with the fixed threshold technique in ablation studies.


The proposed Boltzmann attention sampling shares similarity with several approaches aimed at making transformers effective for handling long sequences ~\cite{child2019generating, beltagy2020longformer}. However, the proposed method is fundamentally different in that the Boltzmann sampling is done prior to the standard full attention, whereas methods for long sequences directly modify the full attention computation. Conceptually, attention-for-long-sequence methods sample to approximate where as the proposed method sample to focus. In computer vision, Deformable DETR \cite{zhudeformable} reduces computation by focusing on a deformable neighbor around a reference point, achieving sparsity but more in a deformable convolution fashion. Most of these works typically rely on fixed patterns, while \ourmethod learns a spatial probability distribution and selects in a stochastic manner.
\section{Method}

The overall architecture of our model is illustrated in Figure \ref{fig:model}. \textit{BoltzFormer} is a transformer decoder that takes in image features from an image encoder (upper left) and text embeddings from a language encoder (lower left). The output is a binary segmentation mask (upper right) corresponding to the text prompt. The core component is the Boltzmann attention sampling block (center) to be described in details.

\subsection{Problem Setup}
We formulate the problem as pixel-wise binary classification (pixel belongs to the text prompt semantics or not). We encode the input image using a backbone followed by a pixel decoder following \cite{cheng2022masked}, resulting in visual features $\mathbf{V}^{(\lambda)} \in \mathbb{R}^{\frac{H}{\lambda} \times \frac{W}{\lambda} \times d}$ with different down-sampling factors $\lambda$. A semantic map $\mathbf{S} \in \mathbb{R}^{H \times W \times d}$ is derived from the multiscale visual features to represent the unified image semantics. The text prompt is encoded into a sequence of embeddings $\mathbf{T} \in \mathbb{R}^{N_T \times d}$. We classify each pixel $(x,y)$ with 

\begin{equation}\label{eq-confidence}
    U_{xy}(\mu) = \text{sigmoid}(\mu^T \mathbf{S}_{xy}),
\end{equation}

where the $\mu \in \mathbb{R}^d$ vector is found by \ourmethod as a function $\mathcal{B}$ of the image and text features
\begin{equation}\label{eq-function}
    \mu = \mathcal{B}(\mathbf{V}, \mathbf{S}, \mathbf{T}).
\end{equation}

Below we show how \ourmethod finds a good $\mu$ vector based on the image and text features.

\subsection{BoltzFormer}
\textit{BoltzFormer} finds the $\mu$ vector by finding a latent query vector $q \in \mathbb{R}^d$, which maps to $\mu$ through a nonlinear MLP transformation
\begin{equation}\label{eq-mlp}
    \mu = \text{MLP}(q).
\end{equation}

Instead of using a single latent query vector $q$, we initiate an ensemble of $m$ queries, forming $\mathbf{Q} = [q^{(1)}, \cdots, q^{(m)}]$. Each query yields its own prediction at the end. The ensemble reduce uncertainty as Boltzmann sampling is a stochastic process, and also communicate among themselves. 

\subsubsection{Text-conditioned prior}
Prior to taking any input from the image, we start the latent query $\mathbf{Q}_0$ by conditioning on the text embedding $\mathbf{T}$:
\begin{equation}\label{eq:condition}
    \mathbf{Q}_0 = \text{FFN}(\text{LayerNorm}([\mathbf{Q}, \mathbf{T}]+\text{SelfAttn}[\mathbf{Q}, \mathbf{T}])),
\end{equation}
where $\mathbf{Q}$ is the learnable initial query matrix. We concatenate $[\mathbf{Q}, \mathbf{T}]$ before performing self-attention, and split out only the $\mathbf{Q}$ part before the feed-forward network layer $\text{FFN}$. The same operation applies to following $[\mathbf{Q}, \mathbf{T}]$ self-attention blocks.

\begin{figure}[t]
    \centering
    \includegraphics[trim={0 9cm 0 0},clip, width=\linewidth]{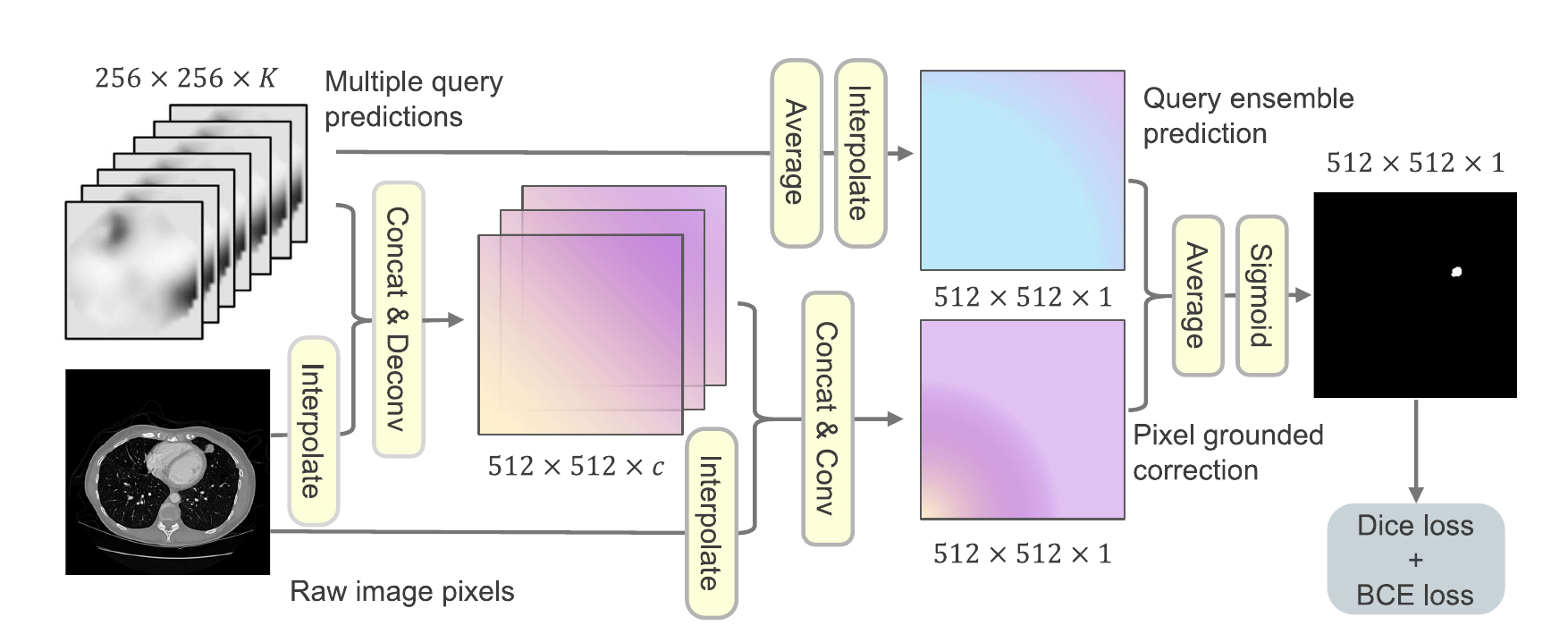}
    \caption{\small The architecture of the PiGMA module. The module takes in the predictions from the final layer queries. The query ensemble prediction part (top row) simply averages the predictions and interpolates to higher resolution. The pixel grounded correction part (de)convolutes the predictions twice into higher resolution. In each convolution layer, we feed in the resized original image and concatenate on the channel dimension. $c$ is the intermediate convolution dimension.
    Finally, the query ensemble prediction and pixel grounded correction are averaged and passed through a sigmoid transformation to produce the pixel-wise probability mask prediction.
    }
    \label{fig:pixel}
\end{figure}

\subsubsection{Boltzmann Attention Sampling}
The latent queries then go through $L$ layers of \ourmethod blocks (big box in the lower center of Fig. \ref{fig:model}). The core in each block is the Boltzmann attention sampling unit (center small box in Fig. \ref{fig:model} and the whole Fig. \ref{fig:attention}) where the queries $\mathbf{Q}_{\ell}$ sample information from the visual features to update themselves. Each Boltzmann attention sampling unit is followed by a self-attention layer for the queries to attend to each other and to the text embeddings. The steps are described in details below.

\paragraph{Step 1. Boltzmann distribution} 
Given the query $q^{(i)}_\ell$ at layer $\ell$, we first use it to compute the Boltzmann distribution on the image, which is a probability field 
\begin{equation}\label{eq-boltz}
    p_{xy}(q^{(i)}_\ell) = \frac{\exp(U_{xy}(\text{MLP}(q^{(i)}_\ell))/\tau_\ell)}{\int_{x' y'} \exp(U_{x'y'}(\text{MLP}(q^{(i)}_\ell))/\tau_\ell)}.
\end{equation}
Here we apply the same MLP transformation in Eq. \eqref{eq-mlp}, and the pixel confidence estimator Eq. \eqref{eq-confidence}. The integral is on all $(x', y')$ on the image field to normalize the distribution. Note that we use a continuous form here to efficiently consider different scales of image features. The discrete distribution can be easily obtained through grid interpolation in real implementation.

Based on Eq. \eqref{eq-boltz}, areas with higher confidence scores $U_{xy}$ are assigned higher sampling probabilities. The sampling temperature $\tau_\ell$ controls the concentration of the distribution in high-confidence areas. As layer $\ell$ increases, we gradually decrease the temperature according to $\tau_\ell = \tau_0/(1 + \ell)$, where $\tau_0$ is the temperature at the base layer $\ell=0$. As illustrated at the bottom of Figure \ref{fig:model}, we encourage exploration across the entire image in the early layers and progressively shift towards exploitation of the most confident regions.

\paragraph{Step 2. Random attention sampling}
We then sample from the visual feature $\mathbf{V}^{(\lambda)}$ according to the Boltzmann distribution $p_{xy}(q^{(i)}_\ell)$ estimated by query $q^{(i)}_\ell$. (In each layer we use one of the visual feature scales $\lambda$. We leave it to the implementation details in appendix.) As $\mathbf{V}^{(\lambda)}$ forms a down-sampled grid on the image, we first interpolate $p_{xy}(q^{(i)}_\ell)$ to assign probability for each patch. We draw from the distribution over the patches for $N$ independent trials\footnote{For computational efficiency, we sample from a Bernoulli distribution with probability $1 - (1 - p(x, y; q^{(i)}_\ell))^N$ for each $(x, y)$ in parallel as an approximation.}. The union of the sampled patches forms an attention set $\mathcal{A}^{(i)}_\ell$. Since one patch can be sampled multiple times, the total size of the sampled set satisfies $|\mathcal{A}^{(i)}_\ell| \leq N$.

Each query vector $q^{(i)}_\ell$ then exclusively attends to the visual features in the sampled region and adds to itself. In the multi-head cross-attention, the attention score on the image for each head $j$ is calculated as

\begin{align*}
\alpha^j_{xy} = \frac{\exp\left( W_j^Q q^{(i)}_\ell \cdot W_j^K \mathbf{V}^{(\lambda)}_{xy} / d\right)}{\sum_{(x',y') \in \mathcal{A}^{(i)}_\ell} \exp \left( W_j^Q q^{(i)}_\ell \cdot W_j^K \mathbf{V}^{(\lambda)}_{x'y'} / d \right)} 
\end{align*}
for $(x,y) \in \mathcal{A}^{(i)}_\ell$, and $0$ elsewhere\footnote{We used the equivalent PyTorch attention masking feature in practice.}. The output of head $j$ attention is
\begin{align*}
H_j  = \sum_{(x',y') \in \mathcal{A}^{(i)}_\ell} \alpha^j_{x'y'} W_j^V \mathbf{V}^{(\lambda)}_{x'y'}.
\end{align*}
The query is then updated by combining the heads and adding to itself:
\begin{align*}
    q^{(i)}_{\ell+1} = q^{(i)}_\ell + [H_1, \cdots, H_h] W^O.
\end{align*}

We apply the same process for all query vectors $q^{(1)}_\ell, \cdots, q^{(m)}_\ell$, followed by a layer normalization on the updated query ensemble before advancing to the next step.

\paragraph{Step 3. Inter-query attention}
After all the queries go through the Boltzmann attention sampling process, we apply a self-attention block such that the information is shared among themselves and with the text embeddings (to the right of the Boltzmann attention sampling box in Fig. \ref{fig:model}):
\begin{equation}
    \text{FFN}(\text{LayerNorm}([\mathbf{Q}_{\ell + 1}, \mathbf{T}_\ell]+\text{SelfAttn}[\mathbf{Q}_{\ell + 1}, \mathbf{T}_\ell])).
\end{equation}
Again, we split out the $\mathbf{Q}$ part before going through the feed-forward network layer. Here $\mathbf{T}_\ell$ is the text embedding from the previous layer. The updated query embeddings and text embeddings are fed into the next \ourmethod block.

\subsection{PiGMA Query Aggregation}
After $L$ blocks of Boltzmann attention sampling, we obtain $m$ finalized latent query vectors $q^{(1)}_L, \dots, q^{(m)}_L$. By applying the MLP transformation in Eq. \eqref{eq-mlp} and estimating pixel-wise confidence score using Eq. \eqref{eq-confidence}, each query yields a mask prediction. In this work we use the PiGMA module illustrated in Fig. \ref{fig:pixel} to aggregate the mask predictions from multiple queries. The aggregated prediction combines the following two components. 

\paragraph{Query Ensemble Prediction} As the $m$ predicted masks are the final pixel-wise confidence estimation, this component is simply the average of the mask predictions:
\begin{equation}
    M = \frac{1}{m} \sum_{i=1}^m \text{MLP}(q^{(i)}_L) \cdot \mathbf{S}.
\end{equation}

\paragraph{Pixel-Grounded Correction} To better address the intrinsic randomness in the query predictions, we apply a lightweight two-layer convolutional network to the mask predictions, grounding on the original image pixels to give a high-resolution correction to the ensemble prediction. We illustrate the architecture in Figure \ref{fig:pixel}.

The final output from PiGMA is the average of these two components. We further apply a pixel-wise sigmoid transformation to obtain a probabilistic mask prediction. The final mask prediction is supervised using the sum of the Dice loss \cite{milletari2016v} and binary cross-entropy loss.

\subsection{Remarks}
We finish this section with a few technical remarks.

\begin{itemize}
    \item The goal of \textit{BoltzFormer} is to reduce the attention region by first ``guessing'' on a learnable probability distribution. Each ``guess'' doesn't need to be perfect, as there are multiple queries and rounds. The more relevant feature in each sampled set will be picked out by the cross-attention block efficiently, and shifts the updated query to focus more on those regions. Compared to standard attention which tries to focus on the desired feature only through optimizing on the massive number of tokens, sampling plus attention relieves the learning burden when the target region is very small.

    \item When \textit{BoltzFormer} advances to the next round, features sampled in the previous round that are more relevant will be amplified through cross-attention. As the attention output is added to the original query vector, the next round’s Boltzmann distribution, estimated by the updated query vector, will more focus on the relevant features. Based on this mechanism, it is beneficial to spread out the sampling distribution in the early layers, reducing the likelihood of omitting any relevant features. The annealing temperature scheduling is designed to balance the exploration and exploitation.

    \item The ensemble of multiple queries not only diversify the exploration in Boltzmann sampling, but also boosts the final performance by their inter-communication through the self-attention layers. Once one of the queries get closer to the desired feature, all other queries can update towards the desired state. 
\end{itemize}

\section{Experiments}

In order to evaluate the effectiveness of \textit{BoltzFormer}, we conducted extensive experiments on seven public image segmentation datasets from Medical Segmentation Decathlon (MSD) \cite{antonelli2022medical}, LIDC-IDRI \cite{armato2011lung} and AMOS22 \cite{ji2022amos}. We used the GPT-4 enriched data from BiomedParse\footnote{We note that the processed variants are different from the original versions in that the images are in 2D and the split definition could be different.}~\cite{zhao2024biomedparse} to evaluate text prompt segmentation for the targets in these datasets. Example prompts include ``right kidney in abdominal MRI'' and ``neoplasm in lung CT''. We ensure each of these datasets contain segmentation object types that are smaller than 1\% of the image size in terms of area. The total benchmark suite covers a wide range of object sizes from below 0.002\% to over 20\% of image area, crossing four orders of magnitude. We provide detailed information in the Appendix.

\subsection{Baselines}
We compared \ourmethod with three categories of baseline implementations.

\noindent \textbf{Segmentation FM decoder comparison: } 
We took the transformer decoder architectures of SOTA promptable image segmentation foundation models (FM), including SAM \cite{kirillov2023segment} (\textit{ICCV} 2023), SEEM \cite{zou2024segment} (\textit{NeurIPS} 2024), and the recent SAM 2 \cite{ravi2024sam} (image version). To compare with \ourmethod in the same text prompt setting, we customized SAM and SAM 2 to take in text embeddings using UniCL \citep{yang2022unified} as the language encoder. For image backbone, we used the Hiera \citep{ryali2023hiera, bolyawindow} model with masked autoencoder (MAE) pre-training \citep{he2022masked}, following the implementation in SAM 2 \citep{ravi2024sam}. We compared the decoders in settings with different backbone configurations (Hiera-S and Hiera-BP). To test the compatibility of \ourmethod with different backbone architectures, we also implemented an additional experiment using the Focal-L \cite{yang2022focal} backbone initialized from SEEM \citealp{zou2024segment}. The pixel decoder architecture follows \cite{cheng2022masked, zou2024segment, zou2023generalized}. For all decoder-backbone combinations, we trained on the entirety of the benchmark suite jointly. 

\noindent \textbf{Pre-trained foundation model comparison: } We took the weights of the state-of-the-art biomedical image foundation model BiomedParse \cite{zhao2024biomedparse} for comparison. BiomedParse takes in text prompt and performs segmentation for objects in medical images. The model was not further fine-tuned exclusively on the benchmark suite in this paper.

\noindent \textbf{Expert model comparison: } We used nnU-Net \cite{isensee2021nnu} as the expert model baseline. We trained 35 nnU-Net models in fully supervised setting performing
binary segmentation for each object type in each dataset, without text prompting.

\begin{figure}[t]
    \centering
    \includegraphics[trim={0 7cm 0 0},clip, width=\linewidth]{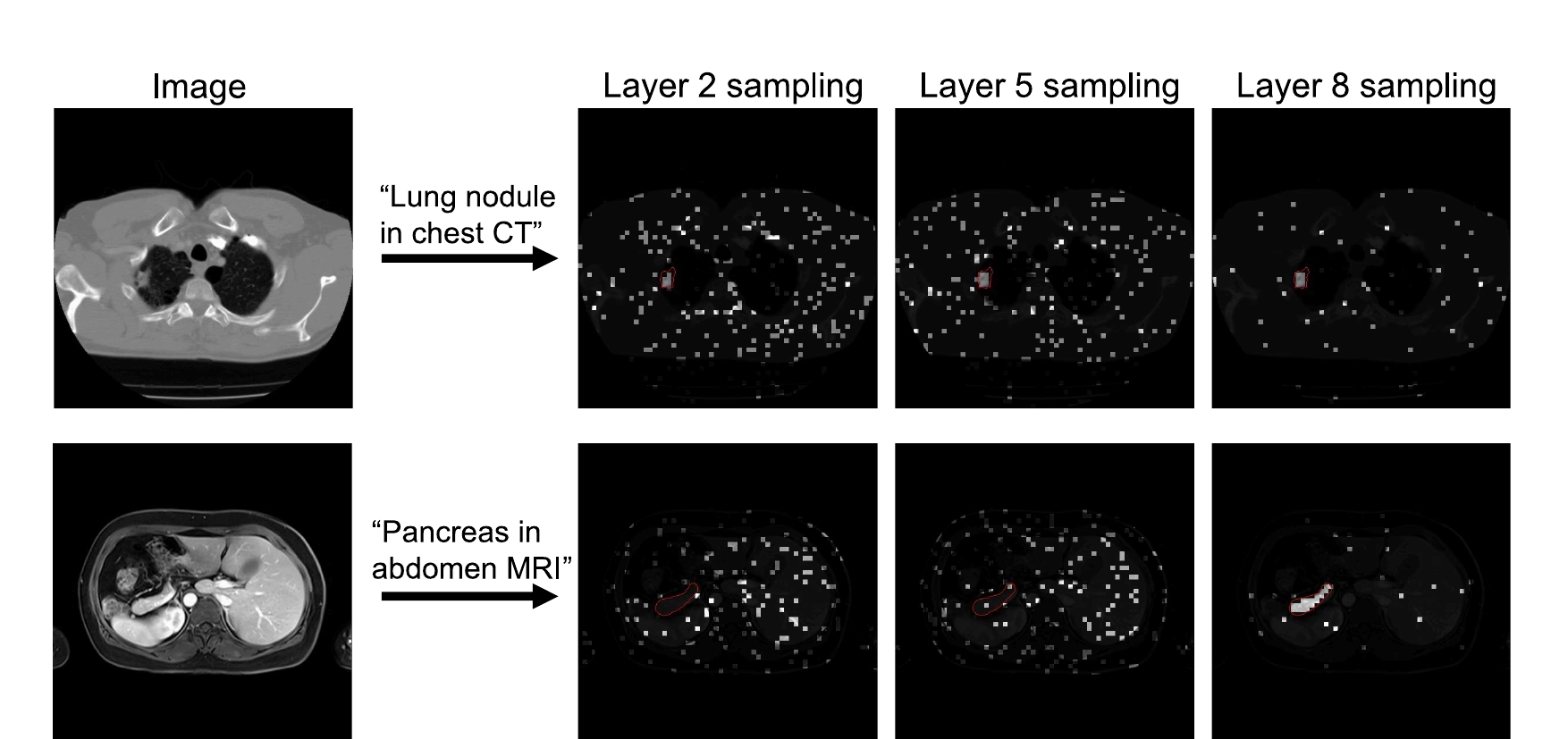}
    \caption{\small Examples of Boltzmann sampling from the intermediate layers during inference. For each image and text prompt, the queries only attend to the samples patches at each layer. The dark region is completely masked out in that layer. Boundaries of the ground truth object is marked in red.}
    \label{fig:demo}
\end{figure}

\begin{table*}[t]
  \centering
  \caption{\small Evaluation results on segmentation benchmarks measured in mean Dice score (\%). The first three sections are baseline models of the three categories. We present results of \ourmethod in the last section. For \ourmethod and the segmentation FM decoder baselines, we present each implementation as ``DECODER+BACKBONE''. We mark each baseline decoders implementation with ``text'' to emphasize that they were all customized to take text prompts using the exact same language encoder as \ourmethod. We mark the highest scoring model with \textbf{bold} in each column. We use \underline{underline} to denote the best model under the same backbone size (e.g, S, BP).}
  \small
  \label{tab:results}
  \begin{tabular}{@{}lcccccccc@{}}
    \toprule
    Method &   Avg. & LIDC & AMOS-CT & AMOS-MRI & MSD-Lung & MSD-Panc. & MSD-HepVes & MSD-Colon \\
    \midrule
    \multicolumn{9}{l}{\it Expert models trained for each class in each dataset separately } \\
    nnU-Net   & 67.3 & 64.8 & 85.0 & 85.2 & 60.2 & 52.4 & 61.3 & 62.4 \\
    \midrule
    \multicolumn{9}{l}{\it Pre-trained biomedical foundation model} \\    
    BiomedParse & 73.0 & 73.8 & 91.9 & 87.6 & 66.1 & 60.2 & 64.8 & 66.4 \\
    \midrule
    \multicolumn{9}{l}{\it Segmentation FM decoders trained on this benchmark suite with custom text encoders and backbones} \\     
    SAM+Heira-S (text)      & 67.0 & 67.1 & 88.4 & 83.9 & 61.6 & 55.1 & 61.2 & 52.0 \\
    SAM+Heira-BP (text)     & 64.9 & 62.8 & 86.7 & 81.2 & 57.2 & 51.4 & 60.8 & 54.0 \\
    SAM2+Heira-S (text)     & 65.6 & 65.4 & 88.2 & 82.6 & 59.8 & 52.8 & 59.8 & 50.6 \\
    SAM2+Heira-BP (text)    & 66.1 & 63.2 & 88.1 & 82.0 & 64.2 & 52.9 & 60.1 & 51.6 \\
    SEEM+Heira-S (text)     & 71.5 & 72.1 & 91.1 & 88.2 & 65.9 & 61.4 & 63.9 & 58.0 \\
    SEEM+Heira-BP (text)    & 73.8 & 72.9 & 92.0 & 89.3 & 69.3 & \underline{\textbf{64.6}} & 65.7 & 63.3 \\
    \midrule
    \multicolumn{9}{l}{\it BoltzFormer (our method) trained on this benchmark suite with custom text encoders and backbones} \\
    BoltzFormer+Heira-S     & \underline{73.8} & \underline{73.3} & \underline{91.3} & \underline{88.7} & \underline{70.4} & \underline{63.7} & \underline{64.4} & \underline{65.2} \\
    BoltzFormer+Heira-BP    & \underline{74.5} & \underline{73.6} & \underline{92.0} & \underline{89.3} & \underline{\textbf{71.4}} & 61.0 & \underline{66.7} & \underline{\textbf{66.7}} \\
    BoltzFormer+FocalL      & \textbf{75.2} & \textbf{75.4} & \textbf{92.7} & \textbf{90.2} & 70.2 & 64.0 & \textbf{67.0} & \textbf{66.7} \\
    \bottomrule
  \end{tabular}
\end{table*}

\subsection{Results}

Table \ref{tab:results} shows the Dice score evaluation of \textit{BoltzFormer} compared with all the baseline models. Holding the image backbone constant, \ourmethod outperformed all other segmentation FM decoders in mean Dice score averaged across the benckmark suite. The average performance improvements against the baseline decoders are: 12.4\% against SAM, 12.6\% against SAM 2, and 2.1\% against SEEM. On individual benchmarks, \ourmethod won over all baselines in most cases. In pre-trained model comparison, \ourmethod outperformed BiomedParse on average and on each individual dataset. \ourmethod also outperformed nnU-Net with 35 task-specific expert models. Lastly, \ourmethod on the FocalL backbone shows dominant performance across the benchmark suite.

In terms of failure analysis, the ratio of failure cases where the model completely missed the target was estimated at 1.4\%. Common failure examples are: 1. extremely small targets (e.g. a few pixels), 2. low contrast, 3. presence of other plausible objects.

\subsection{Performance on small objects} 
To show the advantage of \ourmethod in segmenting small objects, we filtered the segmentation examples into small and large groups with the threshold at $1\%$ of the total image area. Table \ref{tab:small_eval} shows the performance stratified by objects size, where \ourmethod significantly outperformed all other baselines on small objects. Interestingly, we noticed that while \ourmethod was leading on the large object group as well, the gap between \ourmethod and the best competing method on large objects is very small, proving that the improvement from our approach was mainly contributed by the small objects.

\begin{table}[h]
  \centering
  \small
  \caption{\small Dice score (with std) on small objects ($<1\%$ image area), large objects ($\geq 1 \%$ image area), and all objects. We calculated the average Dice score for each size group on each dataset in the benchmark suite, and reported the dataset average weighted by the ratio of the corresponding size group in each dataset.}
  \begin{tabular}{@{}l|cccc@{}}
    \toprule
     & SAM & SAM2 & SEEM & BoltzFormer \\
    \midrule
    Small & 64.5 (0.52)  & 62.1 (0.53) & 68.9 (0.50) & \textbf{71.4 (0.44)} \\
    Large & 82.3 (0.57)  & 82.3 (0.57) & 87.1 (0.36) & \textbf{87.5 (0.35)} \\
    All & 67.1 (0.46)  & 65.0 (0.46) & 71.5 (0.44) & \textbf{73.7 (0.38)} \\
    \bottomrule
  \end{tabular}
  \label{tab:small_eval}
\end{table} 

\subsection{Visualization of Boltzmann attention sampling}

In Fig. \ref{fig:demo} we show the two examples of Boltzmann attention sampling during text prompt segmentation inference. The query only attends to the visual features on the sampled patches. The sampling is spread out at the earlier layers, for the model to explore the image features. In the middle layers, \ourmethod begins to exhaust the features in the target region, while continuing to explore other parts of the image. The sampling is highly concentrated on the target regions when getting close to the last layer to refine the prediction with the most relevant features.

Note that in the second example, the query didn't attend to any feature from the target region until layer 5, however, after that the model quickly focused on the target region.

\section{Ablation studies}

We conducted thorough ablation studies for all newly introduced components in \ourmethod to examine their effectiveness. We used Heiera-S as the backbone, and held all non-ablated parts constant for all experiments. Standard deviation estimates are provided in parentheses. 

\subsection{Boltzmann attention sampling}
To show the effectiveness of Boltzmann attention sampling in attention masking, we compared with the standard full attention, as well as fixed threshold masking following \citep{cheng2022masked, zou2024segment}. Table \ref{tab:boltz_vs_threshold} shows the average Dice score evaluated across the benchmark suite, where Boltzmann attention sampling showed significant advantage over full attention and fixed threshold masking.
\begin{table}[h]
  \centering
  \small
  \caption{\small Average Dice score (\%) when using regular full attention, fixed attention mask threshold, and Boltzmann attention sampling.}
  \begin{tabular}{@{}ccc@{}}
    \toprule
    Full attention & 0.5 threshold & Boltz. sampling (ours) \\
    \midrule
    71.2 (0.43) & 72.4 (0.41) & \textbf{73.7 (0.38)} \\
    \bottomrule
  \end{tabular}
  \label{tab:boltz_vs_threshold}
\end{table}

\subsection{Temperature for sampling}
Temperature is the critical parameter in balancing exploration vs. exploitation and controlling the annealing behavior. In Table \ref{tab:boltz_temp} we experimented with different base temperatures in the layer-decay formula $\tau_\ell = \frac{\tau_0}{\ell + 1}$. We found that a balanced value $\tau_0=1$ yielded the best performance. The performance dropped significantly when the temperature is too high, as the model explored for too long and couldn't exploit the most relevant features.

\begin{table}[h]
  \centering
  \small
  \caption{Average Dice score (\%) when changing the base layer temperature in Boltzmann sampling.}
  \begin{tabular}{@{}lcccc@{}}
    \toprule
    Temp. & $\tau_0=0.25$ & $\tau_0=0.5$ & $\tau_0=1$ & $\tau_0=2$  \\
    \midrule
    Dice & 73.6 (0.40) & 73.0 (0.43) & \textbf{73.7 (0.38)} & 72.2 (0.42) \\
    \bottomrule
  \end{tabular}
  \label{tab:boltz_temp}
\end{table}

\subsection{Sampling size}
We experimented with the number of samples drawn by each query in each layer, measured as a percentage of the total number of visual features $N_v$ in that layer. In Table \ref{tab:boltz_N_sample} we observed that with as few as 5\% features \ourmethod is able to deliver satisfactory performance, and 10\% sampling gave the best performance. Further increasing the sample size didn't help with the model performance.

\begin{table}[h]
  \centering
  \small
  \caption{Average Dice score (\%) when changing the number of independent samples in Boltzmann sampling. $N_v$ refers to the total number of visual features in that layer.}
  \begin{tabular}{@{}lcccc@{}}
    \toprule
    \# sample & $5\% N_v$ & $10\% N_v$ & $20\% N_v$ & $50\% N_v$  \\
    \midrule
    Dice & 72.9 (0.40) & \textbf{73.7 (0.38)} & 72.4 (0.42) & 73.6 (0.40) \\
    \bottomrule
  \end{tabular}
  \label{tab:boltz_N_sample}
\end{table}

\subsection{Number of queries}

One key component in \ourmethod is the design of multiple queries which perform independent distribution estimation and sampling. The self-attention layer after each Boltzmann attention sampling layer communicates the useful features across the ensemble. From Table \ref{tab:boltz_query} we observed that as few as 10 queries is enough for best performance, and there was no notable difference when further increasing the number of queries. When there is no ensemble (only 1 query) the performance dropped notably.

\begin{table}[ht]
  \centering
  \small
  \caption{Average Dice score (\%) when changing the number of queries in \ourmethod. }
  \begin{tabular}{@{}lcccc@{}}
    \toprule
    \# queries & 1 & 10 & 32 & 101  \\
    \midrule
    Dice & 72.8 (0.42) &  \textbf{73.8 (0.38)} & \textbf{73.8 (0.38)} & 73.7 (0.38) \\
    \bottomrule
  \end{tabular}
  \label{tab:boltz_query}
\end{table}

\subsection{Text-conditional query prior}
The self-attention layer before the first Boltzmann attention sampling block helps the queries to initialize based on the semantic from the text prompt. Table \ref{tab:text_condition} illustrates the performance gain from text-conditioning (Eq. \eqref{eq:condition}).

\begin{table}[h]
  \centering
  \small
  \caption{Average Dice score (\%) when using unconditioned query prior v.s. using query prior conditioned on input text.}
  \begin{tabular}{@{}cc@{}}
    \toprule
    Unconditioned prior & Text-conditioned prior \\
    \midrule
    72.3 (0.44) & \textbf{73.7 (0.38)} \\
    \bottomrule
  \end{tabular}
  \label{tab:text_condition}
\end{table}

\subsection{PiGMA module}
The PiGMA module at the end of the model provides a nonlinear aggregation of the different predictions from the multiple queries by grounding on the original image. Table \ref{tab:pixel} shows the improvement from pixel grounded correction.

\begin{table}[h]
  \centering
  \small
  \caption{Average Dice score (\%) when using query ensemble prediction only v.s. adding high resolution pixel grounded correction.}
  \begin{tabular}{@{}cc@{}}
    \toprule
    Query ensemble only & With pixel grounded correction \\
    \midrule
    73.2 (0.43) & \textbf{73.7 (0.38)} \\
    \bottomrule
  \end{tabular}
  \label{tab:pixel}
\end{table}
\section{Conclusion}

Detecting and segmenting small objects remains a significant challenge in holistic image analysis. In this paper, we introduced \ourmethod for detecting and segmenting small objects by learning to dynamically sample sparse areas for focused self-attention using a Boltzmann distribution with an annealing temperature schedule. Effectively, \ourmethod adopts a form of reinforcement learning by maintaining layerwise representation (state) and selecting the sparse subset of the image (action) through Boltzmann sampling (policy). In thorough evaluation on standard image analysis datasets with small objects, \ourmethod substantially outperforms prior best methods in segmentation accuracy, while reducing self-attention computation by an order of magnitude. Future work will focus on enhancing accuracy and efficiency, as well as expanding applications to additional modalities and object types.

{
    \small
    \bibliographystyle{ieeenat_fullname}
    \bibliography{main}
}

\clearpage
\setcounter{page}{1}
\maketitlesupplementary

\appendix

\section{Implementation details}

\subsection{Image encoder}
We used Hiera or FocalNet as the image backbone. Both models outputs four scales of features with strides 4, 8, 16 and 32. The pixel decoder takes the multiscale backbone features to output multiscale visual features of resolution $32\times 32$, $64\times 64$, $128\times 128$ and $256\times 256$. The multiscale visual features convolute to semantic map in resolution $256\times 256$. \textit{BoltzFormer} attends to multiscale visual features of resolution $32\times 32$, $64\times 64$ and $128\times 128$ in loops. We repeated the loop of the three scales in this order for three times, resulting in nine layers in total. The query vectors after the ninth layer was used for prediction.

\subsection{Language encoder}
We used UniCL as the language encoder for text prompts. The context length is 77. We input the full token embedding sequence to \textit{BoltzFormer}.

\subsection{Boltzmann sampling}
In our main experiments, we used a sampling ratio of 10\%, which means at each layer the number of independent trials is 10\% of the total number of visual features of that scale. We used default base temperature $\tau_0 = 1$.

\subsection{Training specifics}
We used 32 NVIDIA A100 80GB GPUs for \textit{BoltzFormer} training. We used batch size $12 \times 32$ for Hiera-S and Hiera-BP backbones, and batch size $8 \times 32$ for Focal-L backbone. During training we split out 20\% from the training set for validation, and monitored the validation loss. We trained with early stopping based on validation loss for a maximum of 40 epochs. We used AdamW~\citep{loshchilov2017decoupled} as the optimizer with equal weighted Dice loss and pixel-wise binary cross-entropy loss. Selected training hyper-parameters based on validation loss for learning rates $2 \times 10^{-5}, 4 \times 10^{-5}, 5 \times 10^{-5}, 1 \times 10^{-4}$, and weight decays $10^{-4}, 10^{-3}, 10^{-2}$. We used learning rate $2 \times 10^{-5}$ and weight decay $10^{-2}$ for training \textit{BoltzFormer}.

\subsection{Data augmentation}
We implemented random transformation for the images and ground truth masks. Each training example had 50\% probability to be transformed. We randomly rotated the image and corresponding ground truth mask by $0^\circ, 90^\circ, 180^\circ, 270^\circ$. We randomly cropped the image and the corresponding ground truth mask by shifting the center by $\pm 10\%$ horizontally and vertically, and scaling by $\pm 10\%$. All random augmentation were uniformly sampled.

\section{Dataset details}
We listed the object size statistics for all the benchmark datasets we used for evaluation in Table \ref{tab:data_stats}. The benchmark suite used in our study covered a wide range of object sizes, from 0.002\% to 20\% of the image area, crossing four orders of magnitudes. The majority of the object types in each dataset have mean object size less than 1\% of the image area.

\begin{table*}
\centering
\caption{Statistics about the benchmark datasets. We listed all the object types for segmentation in each of the datasets, along with the mean, median, min and max of the object size as a ratio to the total image area.}
\label{tab:data_stats}
\begin{tabular}{llccccc}
\toprule
                 Dataset &              Object type &  Area mean (\%) &  Area median (\%) &  Area min (\%) &  Area max (\%) &  Support \\
\midrule
               LIDC-IDRI &              nodule &     0.059 &       0.029 &    0.002 &    0.802 &     1733 \\
         MSD-Lung &               tumor &     0.283 &       0.200 &    0.011 &    1.062 &      242 \\
     MSD-Pancreas &               tumor &     0.629 &       0.295 &    0.018 &    5.106 &      575 \\
     MSD-Pancreas &            pancreas &     0.567 &       0.465 &    0.029 &    2.431 &     1635 \\
MSD-HepaticVessel &               tumor &     0.535 &       0.281 &    0.005 &    3.652 &      918 \\
MSD-HepaticVessel &              vessel &     0.211 &       0.178 &    0.017 &    0.878 &     2204 \\
        MSD-Colon &               tumor &     0.472 &       0.341 &    0.036 &    2.287 &      245 \\
               AMOS22-CT &       adrenal gland &     0.185 &       0.169 &    0.037 &    0.503 &      420 \\
               AMOS22-CT &           esophagus &     0.142 &       0.102 &    0.018 &    1.189 &     1964 \\
               AMOS22-CT &            postcava &     0.237 &       0.218 &    0.034 &    0.751 &     4105 \\
               AMOS22-CT &              spleen &     1.952 &       1.752 &    0.142 &    6.587 &     1587 \\
               AMOS22-CT & right adrenal gland &     0.087 &       0.073 &    0.010 &    0.304 &      571 \\
               AMOS22-CT &         left kidney &     1.194 &       1.206 &    0.065 &    2.649 &     1776 \\
               AMOS22-CT &              kidney &     2.509 &       2.473 &    0.334 &    4.798 &     1465 \\
               AMOS22-CT &            duodenum &     0.497 &       0.425 &    0.036 &    2.816 &     1677 \\
               AMOS22-CT &             bladder &     2.216 &       1.780 &    0.161 &    7.851 &      864 \\
               AMOS22-CT &         gallbladder &     0.664 &       0.490 &    0.035 &    2.946 &      712 \\
               AMOS22-CT &               liver &     7.896 &       7.285 &    0.694 &   21.498 &     4648 \\
               AMOS22-CT &        right kidney &     1.171 &       1.191 &    0.052 &    2.295 &     1649 \\
               AMOS22-CT &  left adrenal gland &     0.091 &       0.082 &    0.004 &    0.279 &      635 \\
               AMOS22-CT &            pancreas &     0.818 &       0.653 &    0.071 &    3.016 &     1345 \\
               AMOS22-CT &             stomach &     3.222 &       2.724 &    0.165 &   11.275 &     1837 \\
               AMOS22-CT &               aorta &     0.433 &       0.281 &    0.051 &    3.957 &     4409 \\
              AMOS22-MRI &       adrenal gland &     0.115 &       0.111 &    0.048 &    0.216 &       49 \\
              AMOS22-MRI &           esophagus &     0.064 &       0.058 &    0.022 &    0.167 &      134 \\
              AMOS22-MRI &            postcava &     0.135 &       0.116 &    0.033 &    0.351 &      463 \\
              AMOS22-MRI &              spleen &     1.353 &       1.296 &    0.130 &    2.707 &      197 \\
              AMOS22-MRI & right adrenal gland &     0.056 &       0.053 &    0.011 &    0.157 &       56 \\
              AMOS22-MRI &         left kidney &     0.917 &       0.955 &    0.151 &    1.774 &      237 \\
              AMOS22-MRI &              kidney &     1.909 &       1.913 &    0.637 &    3.027 &      198 \\
              AMOS22-MRI &            duodenum &     0.280 &       0.264 &    0.075 &    0.860 &      201 \\
              AMOS22-MRI &         gallbladder &     0.407 &       0.366 &    0.046 &    0.914 &       75 \\
              AMOS22-MRI &               liver &     6.060 &       5.950 &    0.891 &   11.907 &      304 \\
              AMOS22-MRI &        right kidney &     0.882 &       0.935 &    0.146 &    1.467 &      224 \\
              AMOS22-MRI &  left adrenal gland &     0.051 &       0.045 &    0.020 &    0.109 &       84 \\
              AMOS22-MRI &            pancreas &     0.583 &       0.522 &    0.106 &    1.626 &      195 \\
              AMOS22-MRI &             stomach &     1.054 &       0.975 &    0.224 &    2.687 &      216 \\
              AMOS22-MRI &               aorta &     0.179 &       0.173 &    0.075 &    0.355 &      490 \\
\bottomrule
\end{tabular}
\end{table*}

\section{Visualizations}
\subsection{Boltzmann sampling}
We visualize the examples of Boltzmann attention sampling through the layers in \textit{BoltzFormer}. In each layer, each query vector only attend to the sampled visual features visualized in the figures. We looped through the three scales of visual features for three times, resulting in nine layers in total. The shaded regions are completely invisible to the query in that layer. We show the Boltzmann sampling for the first query in the ensemble in these examples. 

\begin{figure*}[ht]
    \centering
    \includegraphics[trim={1cm 2cm 1cm 1cm},clip, width=0.9\textwidth]{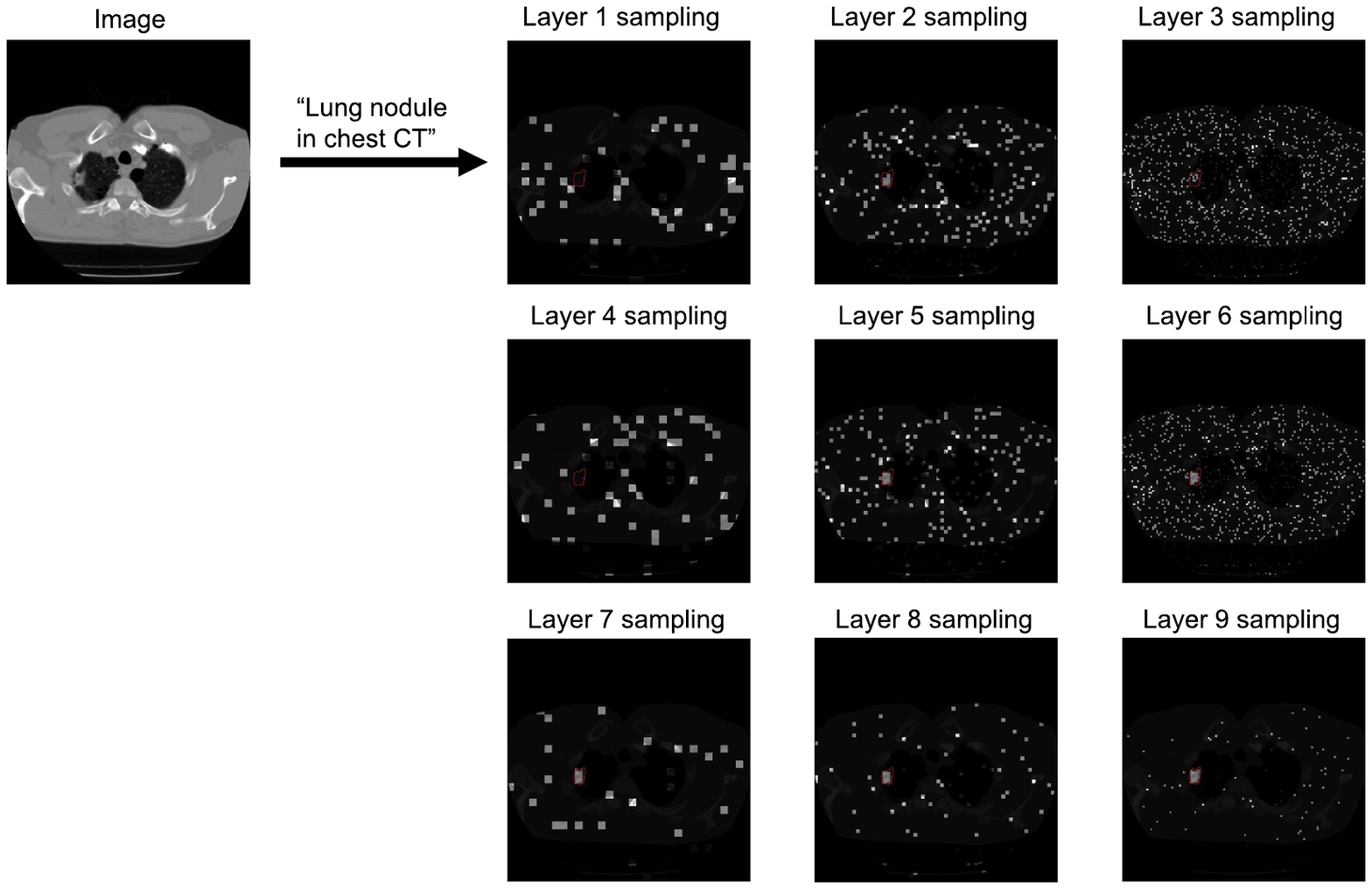}
    \caption{Boltzmann sampling example for lung nodule in chest CT. The sample patches are bright with target region circled in red.}
    \label{fig:sampling_nodule}
\end{figure*}

\begin{figure*}[ht]
    \centering
    \includegraphics[trim={1cm 2cm 1cm 1cm},clip, width=0.9\textwidth]{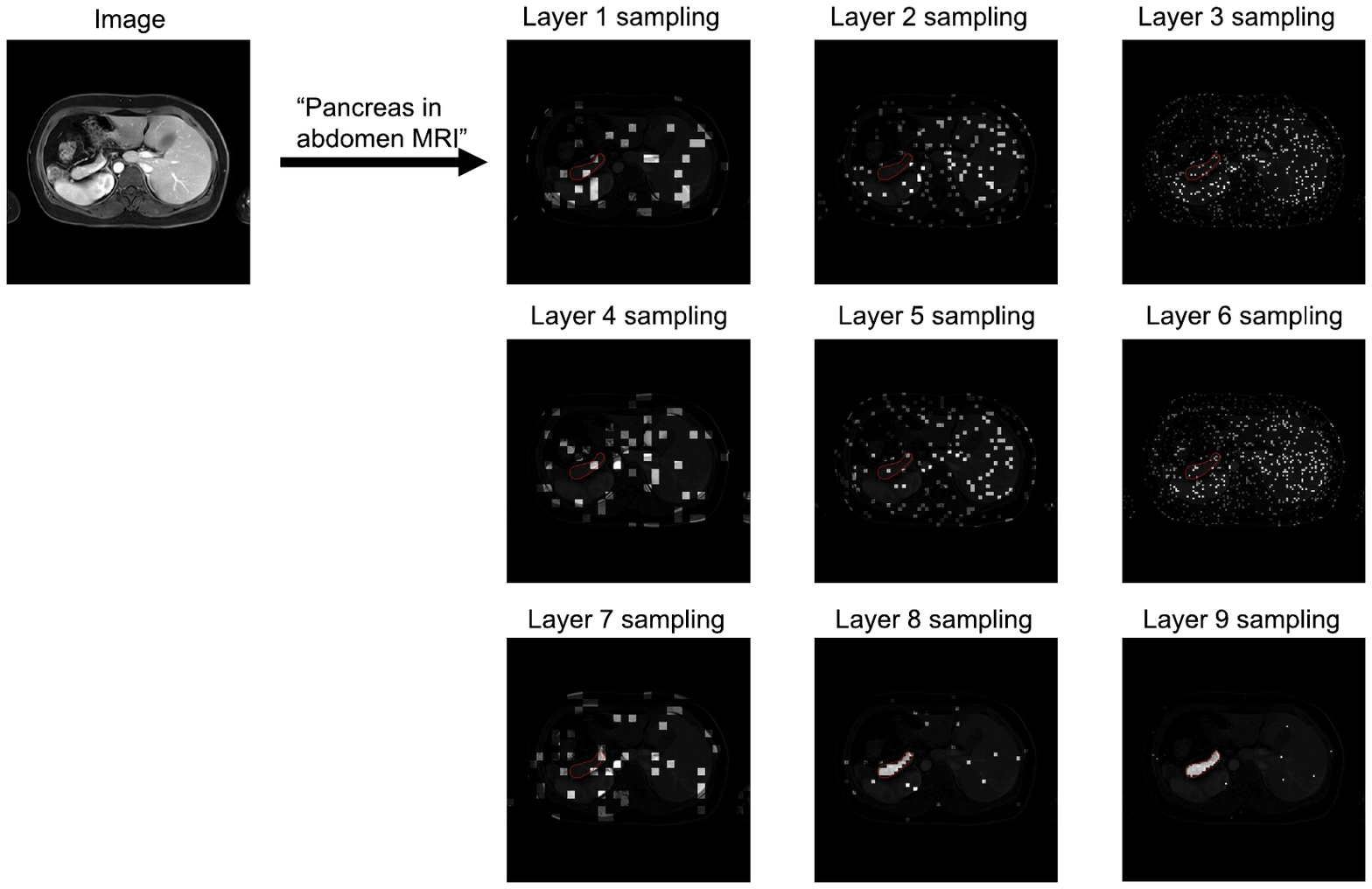}
    \caption{Boltzmann sampling example for pancreas in abdomen MRI. The sample patches are bright with target region circled in red.}
    \label{fig:sampling_pancreas}
\end{figure*}

\begin{figure*}[ht]
    \centering
    \includegraphics[trim={1cm 2cm 1cm 1cm},clip, width=0.9\textwidth]{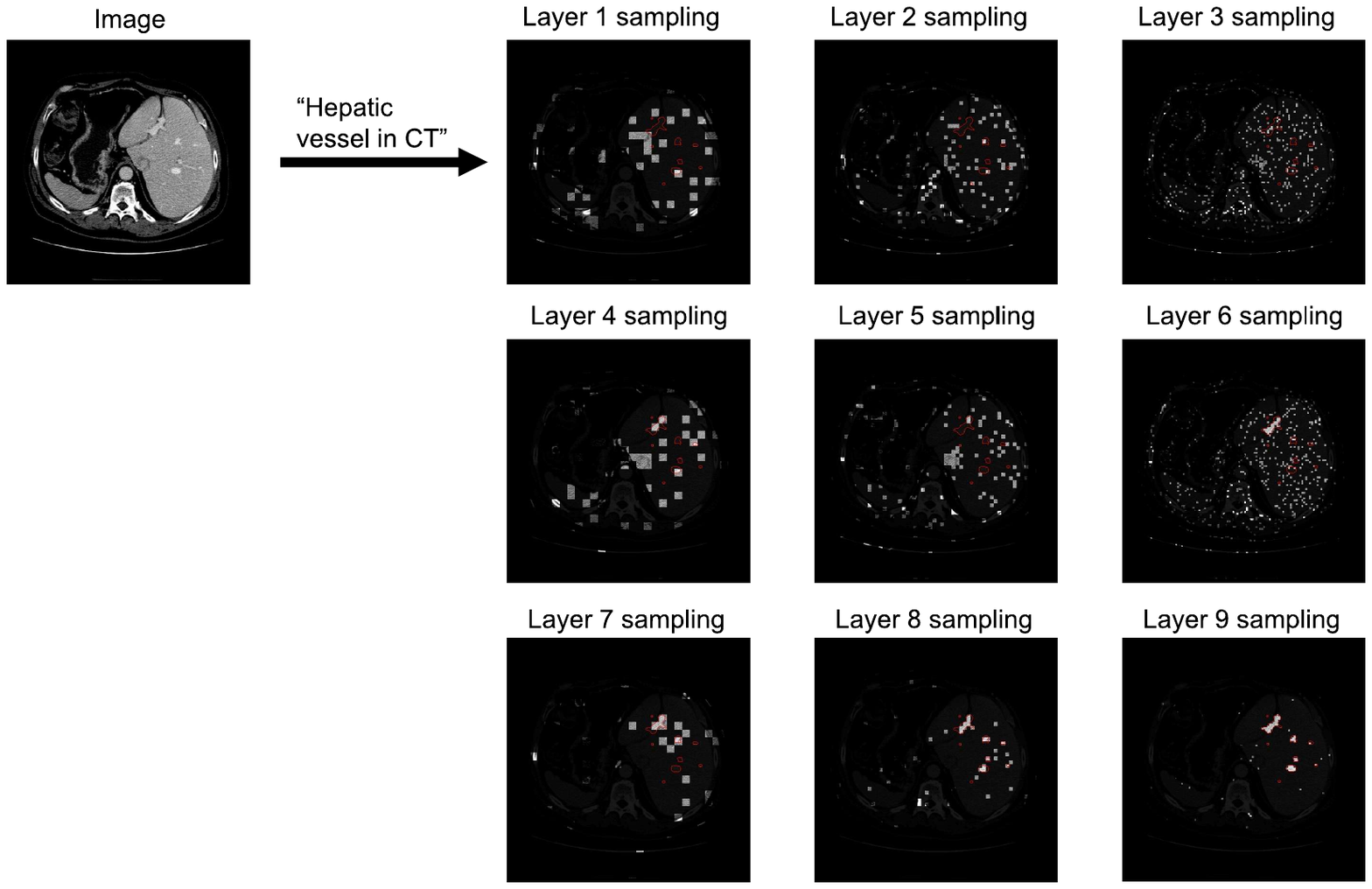}
    \caption{Boltzmann sampling example for hepatic vessel in CT. The sample patches are bright with target region circled in red.}
    \label{fig:sampling_vessel}
\end{figure*}

\begin{figure*}[ht]
    \centering
    \includegraphics[trim={1cm 2cm 1cm 1cm},clip, width=0.9\textwidth]{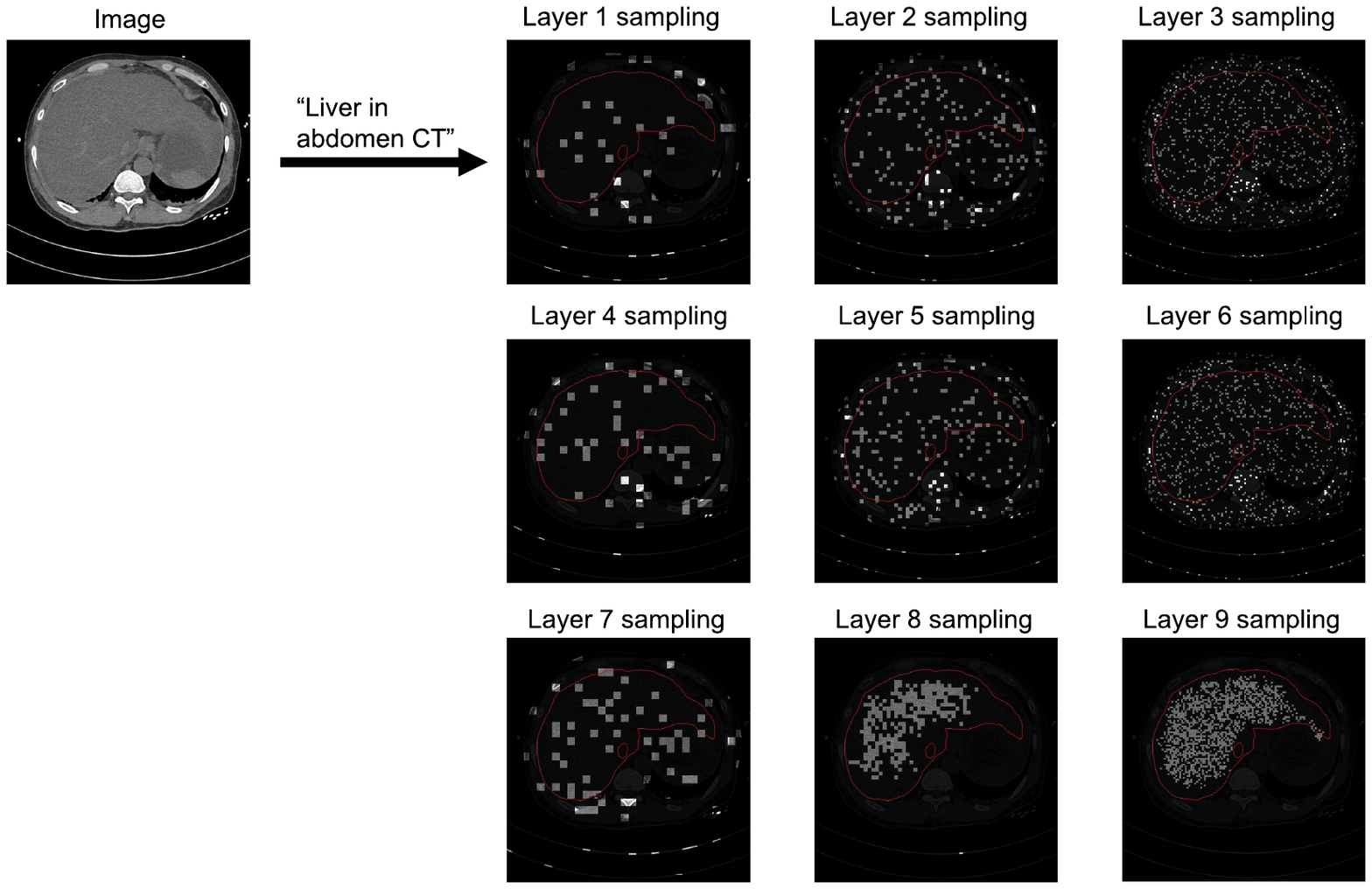}
    \caption{Boltzmann sampling example for liver in abdomen CT. The sample patches are bright with target region circled in red.}
    \label{fig:sampling_liver}
\end{figure*}

From Fig. \ref{fig:sampling_nodule}, \ref{fig:sampling_pancreas}, \ref{fig:sampling_vessel} and \ref{fig:sampling_liver} we can see that the sampling is wide spread during the early layers. The sampling began to concentrate towards the target region during the middle layers, while still exploring the other regions. At the final layers the sampling was highly concentrated on the target regions.

In Fig. \ref{fig:sampling_liver} we show the Boltzmann sampling example for an object with size greater than 20\% of the total image area. Because the sampling ratio is just 10\%, it is impossible to cover all the visual features in the target region. \textit{BoltzFormer} still works for objects with large sizes as the sampled visual features are enough to summarize the semantics of the object.

\subsection{Segmentation result comparison}
\begin{figure*}[ht]
    \centering
    \includegraphics[trim={0 7cm 0 0},clip, width=\textwidth]{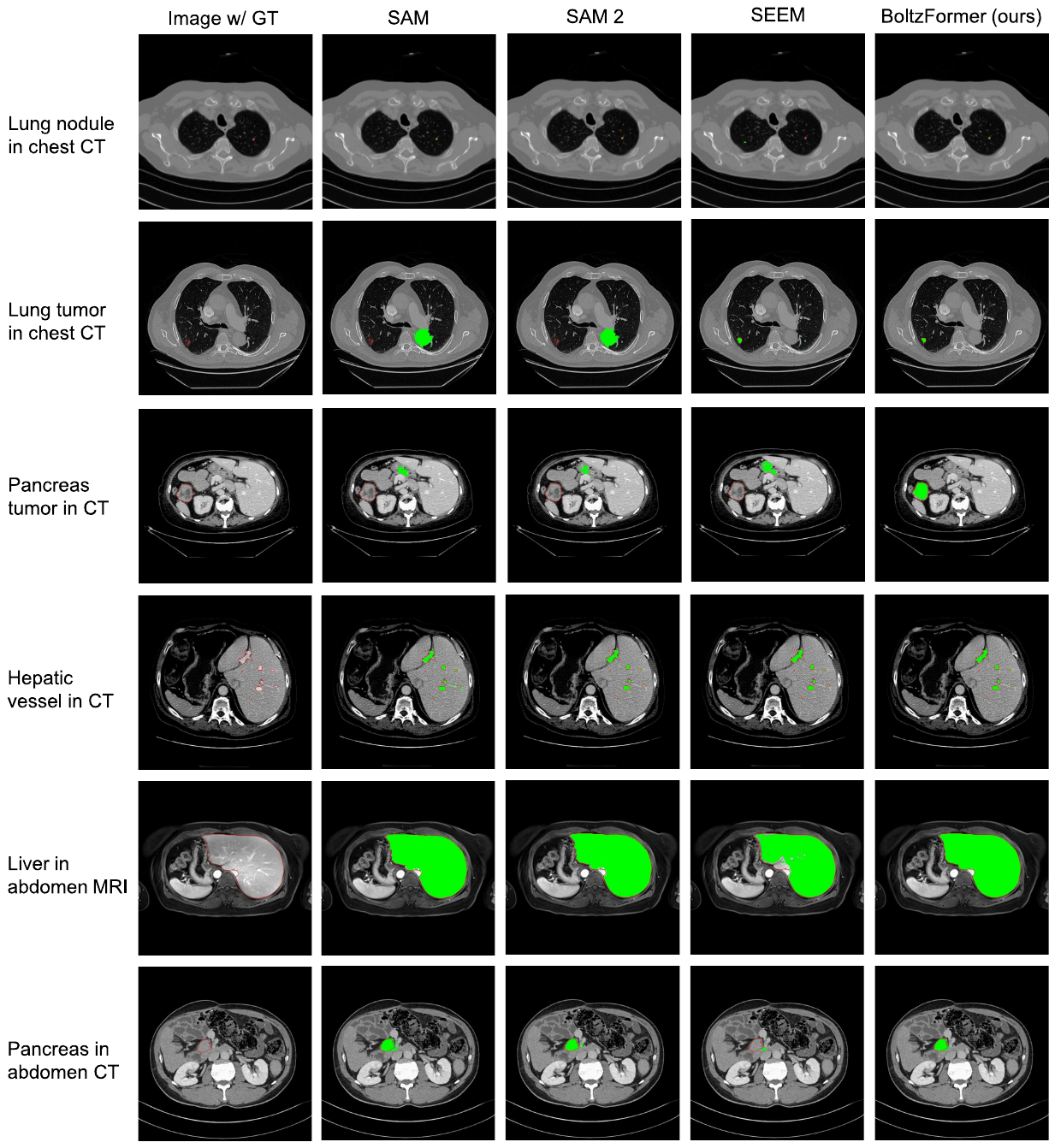}
    \caption{Segmentation prediction examples for baseline decoders and \textit{BoltzFormer}.  Target region is circled in red. Predicted segmentation masks are covered by green. We used the text prompts for all the models. The image backbone is fixed to Hiera-S for all the decoder models.}
    \label{fig:predictions}
\end{figure*}

In Fig. \ref{fig:predictions} we visualize the segmentation results from the baseline models and \textit{BoltzFormer}. All transformer decoders took in text prompts through the UniCL language encoder. The backbone was fixed as Hiera-S for all the models. We visualize the segmantation masks in green, and outline the boundary of the ground truth target in red for reference. 

We can see that \textit{BoltzFormer} consistently delivered accurate segmentation, while the baseline models could miss the target completely when the object is very small (lung nodule and tumor, pancreas tumor). When the object is very large, \textit{BoltzFormer} still output accurate segmentation result.

\end{document}